\documentclass[preprint]{article}

\usepackage[numbers, sort]{natbib}
\usepackage{neurips_2024}
\usepackage{multirow}
\usepackage{graphicx} 

\usepackage[utf8]{inputenc} 
\usepackage[T1]{fontenc}    
\usepackage{hyperref}       
\usepackage{url}            
\usepackage{booktabs}       
\usepackage{amsfonts}       
\usepackage{nicefrac}       
\usepackage{microtype}      
\usepackage{xcolor}         
\usepackage{subcaption}
\usepackage{algorithm}
\usepackage[noend]{algpseudocode}
\usepackage{enumitem}
\usepackage{xcolor}
\newcommand{\model}{DisFormer}
\title{Learning Disentangled Representation in Object-Centric Models for Visual Dynamics Prediction via Transformers}

\author{%
  Sanket Gandhi \\
  Yardi School of AI\\
  IIT Delhi, India\\
  \texttt{sanket.gandhi@scai.iitd.ac.in} \\
  \And
  Atul \\
  Department of Mathematics\\
  IIT Delhi, India\\
  \texttt{mt1210623@maths.iitd.ac.in} \\
  \AND
  Samanyu Mahajan \\
  Department of CSE\\
  IIT Delhi, India\\
  \texttt{cs5190446@cse.iitd.ac.in} \\
  \And
    Vishal Sharma \\
  Department of CSE\\
  IIT Delhi, India\\
  \texttt{vishal.sharma@cse.iitd.ac.in} \\
  \And
    Rushil Gupta \\
  Université de Montréal and Mila\\
  Montréal, Canada\\
  \texttt{rushil.gupta@umontreal.ca} \\
  \And
    Arnab Kumar Mondal \\
  Department of CSE\\
  IIT Delhi, India\\
  \texttt{anz188380@iitd.ac.in} \\
  \And
    Parag Singla \\
  Department of CSE\\
  IIT Delhi, India\\
  \texttt{parags@cse.iitd.ac.in} \\
}

\begin{document}

\maketitle

\begin{abstract}
Recent work has shown that object-centric representations can greatly help improve the accuracy of learning dynamics while also bringing interpretability. In this work, we take this idea one step further, ask the following question: "can learning disentangled representation further improve the accuracy of visual dynamics prediction in object-centric models?" While there has been some attempt to learn such disentangled representations for the case of static images \citep{nsb}, to the best of our knowledge, ours is the first work which tries to do this in a general setting for video, without making any specific assumptions about the kind of attributes that an object might have. The key building block of our architecture is the notion of a {\em block}, where several blocks together constitute an object.  Each block is represented as a linear combination of a given number of learnable concept vectors, which is iteratively refined during the learning process. The blocks in our model are discovered in an unsupervised manner, by attending over object masks, in a style similar to discovery of slots \citep{slot_attention}, for learning a dense object-centric representation. We employ self-attention via transformers over the discovered blocks to predict the next state resulting in discovery of visual dynamics. We perform a series of experiments on several benchmark 2-D, and 3-D datasets demonstrating that our architecture (1) can discover semantically meaningful blocks (2) help improve accuracy of dynamics prediction compared to SOTA object-centric models (3) perform significantly better in OOD setting where the specific attribute combinations are not seen earlier during training. Our experiments highlight the importance discovery of disentangled representation for visual dynamics prediction.

\end{abstract}
\section{Introduction}
\label{sec:intro}
Predicting visual dynamics is an important task for a wide variety of applications including those in Computer Vision and Model-based RL. Recent work has seen a surge of approaches which exploit the power of transformers for predicting future dynamics. It has also been observed that making the model object-aware can help improve the quality of predictions, as well as bring as interpretability. Accordingly, a number of models such as~\cite{gswm,slotformer} have been proposed which explicitly discover object representations, and make use of a GNN~\cite{gswm} or a transformer~\cite{slotformer} to predict future dynamics. Coming from the other side, there has been some recent work on unsupervised discovery of object attributes in static images~\cite{nsb}. To the best of our knowledge, there is no existing work which does this for video, in a general setting, without making an explicit assumption about the kind of attributes that an object might have.

Motivated by this research gap, we ask the following question: Is it possible to learn disentangled representation in an unsupervised manner in object-centric models, and does such a representation help improve the results, for the task of video dynamics prediction. Presumably, such a representation, if learned well, could also help better generalization in OOD settings, i.e., when certain attribute combinations, either appearing in a single object or multiple objects, have not been seen during the training. In response, we propose a novel architecture, which factorizes the object representation using the notion of a {\em block}, where each block can be thought of representing a latent attribute. Blocks in our model are represented via a fixed~\footnote{this is a hyperparameter} number of learnable concept vectors. An easy way to think about concepts is to refer to the idea of a set of basis vectors, such as $RGB$ for the latent color attribute. We note that these are discovered automatically in our model, without any explicit supervision. Once the block representation has been discovered, the objects, factorized in terms of blocks, are passed via a transformer based model, to predict the future state, again represented in terms of blocks at the next time step. 

In terms of overall architecture pipeline, our algorithm proceeds as follows. We first use make use of SAM~\citep{sam}, initialized with with points obtained from another object-centric model, SAVi~\citep{savi}, which represents objects in the form of slots, to discover object masks. A simple CNN encoder is then used to get the object representation from object masks. Once the object latents have been discovered, each of them is attended by a set of blocks, which are iteratively refined, to learn a distentangled representation, as a linear combination of learnable concept vectors. These blocks are then coupled via a self-attention module, to learn an object aware representation, which is then passed to transformer to predict future dynamics. Our decoder is an extension of spatial-broardcast-decoder~\cite{sbd}, extending to the factored block representation, where each block is used to decode a specific dimension of the resultant feature map. Our loss consists of the following components: (a) image reconstruction loss (b) image prediction loss (c) latent block prediction loss (d) object mask prediction loss. While the first loss corresponds to reconstruction at a given time step, the remaining correspond to dynamics prediction at the next time step. Our model is trained jointly, for the discovery of disentangled object representation, as well as the parameters of the dynamics prediction module. We refer to our approach as {\em DisFormer}.

We experiment on a number of 2-D and 3-D benchmark datasets to analyze whether our approach (a) can learn semantically meaningful object representations in terms constituent blocks (b) improve the prediction accuracy of visual dynamics (c) better generalize to OOD settings, where certain attribute combinations have never been before. Our results demonstrate the strength of our model, to disentangle object attributes such as color, size, and position, and significantly help improve the accuracy of visual dynamics prediction, especially in the OOD setting. We also present a series of ablations to give further insights into the working of various components of our model.

Our contributions can be summarized as follows (a) We present the first approach for learning disentangled representation in object-centric models for the task of visual dynamics prediction, without making any explicit assumption about the kind of attributes that an object might have (b) our architecture makes use of a novel object mask detector via SAM initialized with another slot-attention based visual dynamics prediction module (c) we jointly learn the disentangled object representation with dymamics prediction via transformers (d) extensive experiments demonstrate the efficacy of our approach in discovering semantically meaningful object attributes, as well as significantly improve the accuracy of dynamics prediction especially in the OOD setting. We will publicly release the code and datasets on acceptance.

\section{Related Work}
\label{sec:related}
We provide brief over view of related work. Broadly the related work can be split as object-centric models for images, videos, dynamics models and attribute level disentanglement models.  

\noindent
\textbf{Object Centric Image and Video models}
Unsupervised object centric learning from images and videos have been well studied area. Starting from spatial attention modesl AIR~\citep{AIR}, MONet~\citep{MONet}, SPACE~\citep{SPACE}, these models include recent spatial mixture models IODINE~\citep{IODINE}, GENESIS~\citep{GENESIS}, ~\cite{slot_attention} GENESIS-V2~\citep{GENESIS-V2}, SLATE~\citep{SLATE}, and are trained for image reconstruction objectives. 

\noindent 
   \textbf{Dynamics Prediction Models:} 
    Recently, \citep{wu2015galileo,fragkiadaki2016learning,battaglia2016interaction,chang2016compositional} focus on generating predictions of future video sequences, specifically emphasizing objects within the scene. In this context, `object-centric' means that the models and algorithms used for video prediction are designed to understand and track individual objects or entities within the video frames. Object-centric video prediction aims to anticipate how these objects will move, change, or interact in future video frames. However, during training, these approaches require supervision in terms of ground truth properties of the objects. More recent works \citep{ye2019compositional,li2019propagation,qi2020learning,yu2022modular} overcome this limitation of direct supervision by employing a pre-trained detector as object feature extractor and model the dynamics using graph neural network \citep{GCN,GAT,GraphSAGE,GIN}. On the other hand, works such as \citep{van2018relational,kossen2019structured,lin2020improving,wu2023slotformer} develop a completely unsupervised technique. \citep{savi,kabra2021simone,STEVE} decompose videos into temporally aligned slots. 

\noindent 
    \textbf{Object Disentangelment Models:}
    \citep{Jiang*2020SCALOR} proposes a framework to learn scalable object-oriented representation of videos by decomposing foreground and background. \citep{lin2020improving} disentangles various object features such as positions, depth, and semantic attributes separately. \citep{van2018relational,veerapaneni2020entity,creswell2021unsupervised,zoran2021parts,wu2021generative,wu2023slotformer} leverage the prowess of transformer \citep{vaswani2017attention} and GNN \citep{GCN,GAT,GIN} for object aware video representation learning. However, most of the methods don't consider or partially consider the object attributes while modeling the object dynamics. We show that modeling object dynamics with object-tied attributes improves model predictions and enhances generalization capabilities in the context of out-of-distribution (OOD) scenarios.


\section{DISFORMER}
\label{sec:theory}
We describe the architecture of \model\ in this section. There are four important parts to the architecture: (1) Mask Extractor: This module extracts the object masks in an unsupervised manner using pre-trained architectures. (2) Block Extractor: This a novel module that disentangles each object in terms of an underlying {\em block} representation via iterative refinement (3) Dynamics Predictor: This module predicts the next state of object latents, in terms of their block representation via a transformer-based architecture  (4) Decoder: This module inputs the disentangled object representation and decodes it to get the final image. Figure describes the overall architecture. We next describe each module in detail. 
\begin{figure*}[t]
      \centering
      \includegraphics[width=\linewidth]{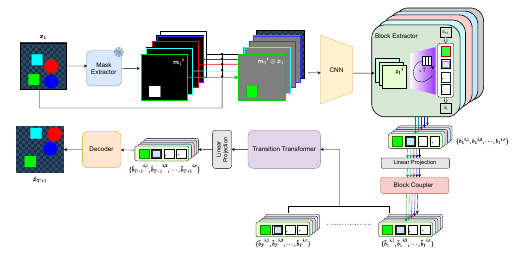}
    \caption{\label{fig:disformer_model} Main architecture for \model{}. SAM and SaVI are the Mask Extractor, and the product operation is the Hadamard Product between Object Masks and the input images. Block Extractor module, takes in the object representations $z_{t}^{i}$, one at a time, along with concept vectors $C$ and outputs the set of block representations for each object $i$. Note that each block has its own set of concept vectors; 
    Attn is the simple dot-product attention module where the input block-based representation of all the objects is converted into a linear combination of the corresponding concept vectors.\ref{sec:theory}}
\end{figure*}
\begin{algorithm}
\small{
\caption{Block Extractor: Inputs are: object features $z_{t}^{i} \in \mathbb{R}^{ f \times d_f }$. Model parameters are: $W_K \in \mathbb{R}^{d_{f} \times d}$, $W_Q \in \mathbb{R}^{ d_{b} \times d}$, $W_V \in \mathbb{R}^{d_f \times d_b}$, concept vectors $C_{b} \in \mathbb{R}^{k \times d_{b}}$ for $b \in \{1,..,r\}$, MLP, GRU}
\label{alg:attr_extract}
\begin{algorithmic}[1]
\State $b_{t}^{i,j}$ = $f_{trans}(b_{t-1}^{i,j})$ \ \ $\forall \ \ j \in \{1,...,r\}$
\For {$t = 1$ to $T$}
\State $b_{t}^{i,j}$ = LayerNorm($b_{t}^{i,j}$) \ \ $\forall \ \ j \in \{1,...,r\}$
\State $A$ = Softmax(($ \frac{1}{\sqrt{d}}  ( b_{t}^{i}W_{Q}) (z_{t}^{i}W_K)^T$, axis = 'block')
\State $A$ = $A$.normalize(axis='feature')
\State $U$ = $A  (z_{t}^{i}W_{V})$
\State $U$ = GRU(state = $b_{t}^{i}$, input = $U$)
\State $U$ = $U$ + MLP($U$)
\For {$j = 1$ to $r$}
\State    $w_{t}^{i,j}$ = Softmax($\frac{1}{\sqrt{d_{j}}}{C_{j} [U[j:]]^T}$)
\State    $b_{t}^{i,j}$ = $C_{j}^T w_{t}^{i,j}$
\EndFor

\EndFor
\State \textbf{return} $s_{t}^{i}$

\end{algorithmic}
}
\end{algorithm}

\subsection{Mask Extractor} \label{ObjectExtractor}
We first pre-train the object-centric model for video which is  modified SAVi\citep{slotformer}. We train this object-centric model with $N$ slots using reconstruction loss on clips from the target dataset. The spatial mixture decoder of the modified SAVi produces temporally aligned object masks while decoding the slots. These object masks are good indicator of where the object is in the frame. To obtain better masks we use SAM \citep{sam} with the prompts generated from the trained SAVi masks. 
Formally given a sequence of frames $\{x_{t}\}_{t = 1}^T$ for $T$ time-steps where $x_{t} \in \mathbb{R}^{H \times W \times 3}$, we first extract the decoder masks from SAVi denoted as $\tilde{m}_{t}^{i} \in [0,1]^{H \times W}$ where $i \in {1,\cdots, N}$ for each frame $x_{t}$. We then threshold the object mask $\tilde{m}_{t}^{i}$ to get binary mask of object $M_{t}^{i} \in \{0,1\}^{H \times W}$. We convolve $M_{t}^{i}$ with all one filter repeatedly for three times. Then from this convolved mask, a subset of pixel positions belonging to maximum pixel value are sampled randomly. Denote this point set as $p_{t}^i$. These pixel positions are considered as positive point prompts to SAM. Positive point prompts from the rest of the objects are chosen as the negative points denoted as $\bar{p}_{t}^i = \cup_{j,j\neq i}p_{t}^j$. These point prompts $p_{t}^i \cup \bar{p}_{t}^i$, mask prompt $m_{t}^i$ and corresponding image $x_{t}$ are forwarded to SAM to produce final mask denoted as $m_{t}^{i}$. These masks are temporally aligned as the prompts were also temporally aligned. 




\subsection{Block Extractor} \label{BlockExtractor}
In our disentangled representation, each object is represented by a set of blocks. We obtain blocks by iterative refinement with attention over a latent object representation. Further, to learn a disentangled representation, we enforce that each block representation is represented as a linear combination of a fixed set of underlying learnable concepts. This combination is discovered at every step of the block-refinement algorithm.
Additionally we initialize the blocks of the object with some learnable function of the object blocks from previous time-step.
\\
Formally the extracted object masks $m_{t}^{i}$ are multiplied element-wise with the corresponding input frame and passed through a  Convolutional Neural Network (CNN) to obtain latent object representations denoted by $z_{t}^{i} \in \mathbb{R}^{f \times d_f}$
Given a latent object representation $z_{t}^{i}$ for the $i^{th}$ object, its representation in terms of blocks is given as $\{b_{t}^{i,j}\}_{j=1}^{r}$, where $r$ is the number of blocks, and is a hyperparameter of the model. We let each $b_{t}^{i,j} \in \mathbb{R}^{d_{b}}$. 
Algorithm~\ref{alg:attr_extract} outlines the iterative refinement step to obtain the block representation.
Each $b_{t}^{i,j}$ is initialized with $f_{trans}(b_{t-1}^{i,j})$ to temporally align the blocks and $f_{trans}$ is implemented as MLP. After an initial layer normalization, we make the block vectors attend over object features to compute the attention scores. These are first normalized over queries (blocks), followed by normalization over keys (object features). The resultant linear combination of object features is first passed through a GRU, followed by an MLP to get the block representation. Finally, unique to our approach, we project each resultant block vector onto the learnable concept space and update its representation as a linear combination of the concepts via projection weights (lines 9 - 12 in Algorithm~\ref{alg:attr_extract}). This step results in discovering the disentangled representation central to our approach. We note that the above iterative refinement steps closely follows the description in \citep{slot_attention}, with the image latent replaced by object latent and slots replaced by blocks depicting factored (disentangled) object representation. 

Recently, ~\cite{nsb} proposed a similar-looking idea of learning disentangled representation for objects as a linear combination of concepts using an iterative refinement over slots, albeit for static images. Our approach is inspired by their work but has some important differences. In their case, the slots still represent objects as in the original slot-attention paper, making the disentanglement closely tied to the learning of object representations. This also means that their approach is limited by a specific choice of object extractor. In contrast, since we work with object representations directly, our approach is oblivious to the choice of the extractor. Further, as already pointed out, their work is limited to static images, whereas we would like to learn these representations for dynamics prediction.

\subsection{Dynamics} \label{Dynamics}
Transformers\citep{transformer} have been very effective at sequence-to-sequence prediction tasks. Some recent work on unsupervised video dynamics prediction \citep{slotformer},\citep{slotdiffusion}, has given SOTA results on this task on 3-D datasets, and have been shown to outperform more traditional GNN based models for this task~\citep{gswm}. While some of these models are object-centric, they do not exploit disentanglement over object representation. Our goal in this section is to integrate our disentanglement pipeline described in Section~\ref{BlockExtractor} with downstream dynamics prediction via transformers. Here are some key ideas that go into developing our transformer for this task:
\begin{enumerate}[leftmargin=*]
\item We linearly project each $b_t^{i,j}$ to a higher $\hat{d}$ dimensional space using a shared $W_{up}$.  $b_t^{i,j} =  W_{up} b_t^{i,j}$, where $W_{up} \in \mathbb{R}^{d_b \times \hat{d}}$.
\item The input to transformer encoder is $T$ step history of all object blocks that is, at time step $t$, the input is $\{b_{t - \bar{t}}^{i,j} | \bar{t} \in \{0,..,T-1\}, i \in \{1,..,N\}, j \in \{1,..,r\}\}$.
\item We need positional encoding to distinguish between (a) different time steps in the history (b) blocks belonging to different objects (c) different blocks in an object. We would also like to maintain the permutation equivariance among objects. Having explicit positional encoding for each object will break the permutation equivariance among objects. Inspired by \citep{nsb} we first add the learnable position encoding $P_j$ to each block and then couple the blocks belonging to same objects using \textit{block coupler} which is one layer transformer encoder. These removes the need of positional encoding for the objects. Similar to \cite{slotformer}, we add sinusoidal time step embedding $P_t$ for each block depending on the frame (time-step) it belongs to. Formally these steps are summarized as: 
\[
\bar{b}_{t}^{i,j} = b_{t}^{i,j} + P_j \ \ \ \ \ \ \ \ \ \ \ \ \ \ 
\tilde{b}_{t}^{i,j} = \texttt{BlockCoupler}(\bar{b}_{t}^{i,1},..,\bar{b}_{t}^{i,r}) \ \ \ \ \ \ \ \ \ \ \ \ \ \ 
\dot{b}_{t}^{i,j} = \tilde{b}_{t}^{i,j} + P_{t}
\]
Here, $\bar{b}$, $\tilde{b}$ and $\dot{b}$ denote the block representations after incorporating block positional embedding, coupling, and frame positional embeddings, respectively.
\item   The transformer encoder output be $\hat{b}_{t}^{i,j}$. We down project it to $d_{b}$ dimensional space using shared $W_{down}$. Thus the final output of dynamics predictor is $\hat{b}_{t}^{i,j} = W_{down}\hat{b}_{t}^{i,j}$
\end{enumerate} 

\subsection{Decoder}
Similar to much of the previous work, we use spatial mixture models to generate the final image. As each object is represented by $r$ vectors of blocks, we use a slightly different approach than previous methods which have a single vector representing the object. We first use shared Spatial Broadcast decoder \citep{sbd} to generate 2D maps $q_{t}^{i,b} \in \mathbb{R}^{f' \times I \times I} $ representing $f'$ features each of size $I \times I$,
corresponding to $b_{t}^{i,j}$. We concatenate these block specific 2D maps to form ${\bf q}_{t}^{i} \in \mathbb{R}^{r f'\times I \times I}$. A CNN is applied on ${\bf q}_{t}^{i}$ to generate final object mask which is normalized across the objects $\hat{m}_{t}^{i}$ and object content $\hat{c}_{t}^{i}$. Final image is obtained as $\hat{x}_{t} = \sum_{i=1}^{N} \hat{m}_{t}^{i} \cdot \hat{c}_{t}^{i} $. Here, we denote the object contents and masks given by the decoder for objects with predicted latent $\hat{b}_{t}^{i}$ as, $\hat{\hat{c}}_{t}^{i}$ and $\hat{\hat{m}}_{t}^{i}$. The resultant image using these masks and contents is denoted as $\hat{\hat{x}}_{t}$.

\subsection{Training and Loss}\label{subsec:training_loss}
We use a history of length $T$ and do a $T'$ step future prediction. We use three-phase training. In first phase we only train \textit{block extractor} and \textit{decoder} with image reconstruction objective. In second Phase we train \textit{dynamics predictor} with \textit{block extractor} and \textit{decoder} being freeze. In third phase, all modules are trained together. \\
{\bf Phase 1:} In first phase we only reconstruct the $T$ history frames. Reconstructing image only involves \textit{block extractor} and \textit{decoder} in forward pass. The reconstruction loss is $\mathcal{L}_{1} = \mathcal{L}_{rec} = \sum_{t=1}^{T} (\hat{x}_{t} - x_{t})^2$ \\
{\bf Phase 2:} In second phase the losses used to train \textit{dynamics predictor} are image prediction loss $\mathcal{L}_{img} = \sum_{t=T+1}^{T'} (\hat{\hat{x}}_{t} - x_{t})^2$, the latent block prediction loss $\mathcal{L}_{block} = \sum_{t=T+1}^{T+T'} \sum_{i,j} (\hat{b}_{t}^{i,j} - {b}_{t}^{i,j})^2$, and mask prediction loss $ \mathcal{L}_{mask} = \sum_{t=T+1}^{T'} \sum_{i}(\hat{\hat{m}}_{t}^i - m_{t}^i)^2$. The total loss is $\mathcal{L}_2 = \mathcal{L}_{img} +\lambda_{block}\mathcal{L}_{block}  + \lambda_{mask}\mathcal{L}_{mask} $.\\
{\bf Phase 3:} In third phase all modules are trained with objective $\mathcal{L}_3 = \mathcal{L}_1 + \mathcal{L}_2$. Decoder parameters are trained only using $\mathcal{L}_{rec}$. 
\noindent


\section{Experiments}
\label{sec:expt}
We perform a series of experiments to answer the following questions: (1) Does \model{} result in better visual predictions than the existing SOTA baselines on the standard datasets for this problem? (2) Does learning disentangled representations with \model{} lead to better performance in OOD attribute combinations? and (3) Can \model{} indeed discover the disentangled representation corresponding to natural object features such as color, size, shape, position, etc.?
We first describe our experimental set-up, including the details of our datasets and experimental methodology, followed by our experimental results, answering each of the above questions in turn.
\subsection{Experimental Setup}
\label{sec:expt:setup}
\subsubsection{Datasets}
We experiment on a total of four datasets for video dynamics prediction; two are 2-dimensional and two are 3-dimensional. While the 2D dataset are toy in nature, they allow us with controlled experiments. 3D domains contain more realistic video dynamics. 

\noindent
\textbf{2D Bouncing Circles (2D-BC):} This 2D dataset is adapted from the bouncing balls Interaction environment in \cite{gswm} with modifications in the number and size of the balls. Our environment comprises three circles of the same size but different colors that move freely in the 2D space with a black background, colliding elastically with the frame walls and each other. The training split contain 1000 videos whereas val and test split contain 300 videos each. All videos have 100 frames. 

\noindent
\textbf{2D Bouncing Shapes (2D-BS):} We create another 2D dataset, which is an extension of the 2D-BC environment with increased visual and dynamics complexity. 
Two circles and two squares move freely in the 2D space with a checkered pattern as background. 
Collisions happen elastically among various objects and frame walls while also respecting respective object geometries.
We use the MuJoCo physics engine~\citep{mujoco} to simulate the domain with camera at the top, and objects with minimum height to have a 2-D environment. The split details are same as that of 2D-BS.

\noindent
\textbf{OBJ3D}: A 3D environment used in GSWM~\citep{gswm} and SlotFormer~\cite{slotformer}, where a typical video has a rubber sphere that enters the frame from front and collides with other still objects. The dataset contain 2912 train videos and 200 test videos. Each video has 100 frames. 

\noindent
\textbf{CLEVRER}: This is standard 3D dataset \citep{CLEVRER:} is similar to OBJ3D with smaller objects. All type of objects can enter the frame from all sides. The dataset contain 10000 train videos and 5000 test videos. Each video has 128 frames. 


    
\subsubsection{Baselines, Metrics and Experimental Methodology}
\textbf{Baselines:} We compare \model{} with two existing baselines, GSWM~\citep{gswm} and SlotFormer~\citep{slotformer}. GSWM is the object-centric generative model. SlotFormer is state-of-the-art slot based object-centric dynamics model. In addition, we also create our own baseline, called DenseFormer, which is variation of our model where objects have dense representation rather than representation factorized in terms of blocks. This is achieved by replacing the block extractor and block coupler modules by a single layer MLP .
This baseline was created to understand the impact of having an object representation factorized in terms of blocks in \model{}. 

\textbf{Evaluation Metrics:} Following existing literature, for 2D-BC and 2D-BS domains, we compare the quality of predicted future frames vis-a-vis the ground truth using two metrics: (1) PErr (Pixel mean-squared error) (2) PSNR. For OBJ3D and CLEVRER, we compare various algorithms on 3 different metrics: (1) PSNR (2) LPIPS \citep{lpips}, and (3) SSIM \citep{SSIM}.

\noindent
\textbf{Experimental Methodology:} We use the official implementation for each of our baselines: \footnote{ GSWM: \href{https://github.com/zhixuan-lin/G-SWM}{https://github.com/zhixuan-lin/G-SWM}, SlotFormer: \href{https://github.com/pairlab/SlotFormer}{https://github.com/pairlab/SlotFormer}}. All the models (except GSWM) are trained to predict 10 frames of video given 6 frames of history. For fair comparison, GSWM, which is a generative model, is provided with 16 frame clips during its training. While testing all the models are conditioned on 6 frames of history and are required to predict future frames. For BC and BS datasets, we unroll for 15 future steps during testing. For OBJ3D we only use only first 50 frames for training and testing as most of interactions between objects ends in first 50 frames \citep{slotformer}\citep{gswm}. For CLEVRER we subsample the frames by a factor of 2 and remove the clips where new objects enter the frame during rollout, as done in existing literature. During testing, OBJ3D and CLEVRER models are unrolled upto 44 and 42 frames respectively following~\cite{slotformer,gswm}, for a direct comparison. For each of the datasets, the reported numbers are averaged over the unrolled future step prediction. 

\noindent
\textbf{OOD Setting:} For OOD experiments we considered the unseen attribute combinations in the three of the datasets. For 2D-BC dataset, all three balls in the video have unique color among red, blue green. The OOD 2D-BC dataset have all possible color combinations of balls. 2D-BS dataset have 2 squares and 2 circles in video clips. For OOD dataset, we created 2D-BS variant where all possible shape combinations are allowed (and were randomly chosen). For OBJ3D, we created OOD dataset where entering sphere is metallic instead of rubber (all training set instances had entering rubber sphere). In each case, we used the model trained on the original dataset, and tested on the OOD dataset. The number of test instances, the history length, as well as the number of rollouts for prediction were kept the same as in the case of in distribution setting for each of the datasets.  



\subsection{Visual Dynamics Prediction}
\begin{figure*}[t]
\begin{subfigure}[t]{0.3\textwidth}
    \centering
      \includegraphics[width=5cm]{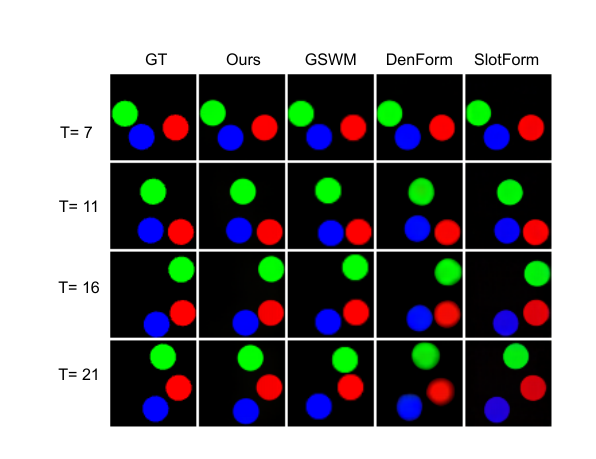}
    \caption{\label{fig:bc_1} 2D-BC }
\end{subfigure}
~
\begin{subfigure}[t]{0.3\textwidth}
    \centering
      \includegraphics[width=5cm]{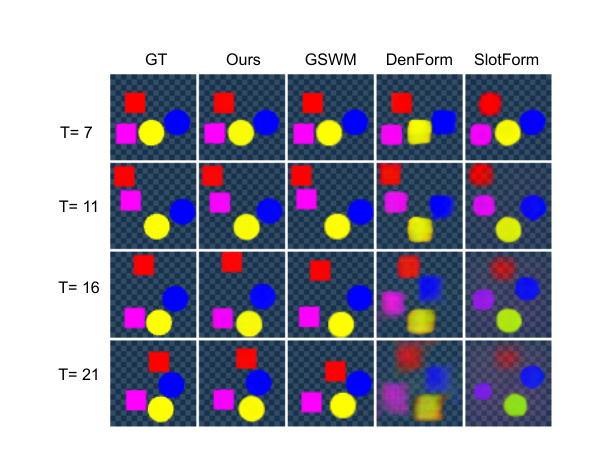}
    \caption{\label{fig:bs_1} 2D-BS }
\end{subfigure}
~
\begin{subfigure}[t]{0.3\textwidth}
    \centering
      \includegraphics[width=5cm]{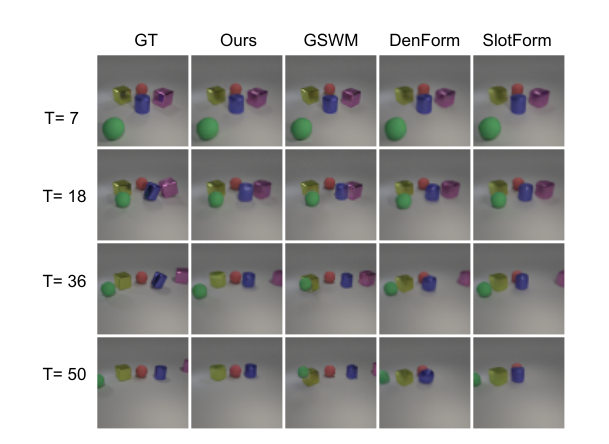}
    \caption{\label{fig:obj3d_1}OBJ3D }
\end{subfigure}
\caption{Rollouts at various time steps for three datasets (in distribution). GT: Ground Truth.}
\label{fig:rollout-indist}
\end{figure*}

\begin{figure*}[h]
\begin{subfigure}[t]{0.3\textwidth}
    \centering
      \includegraphics[width=4.5cm]{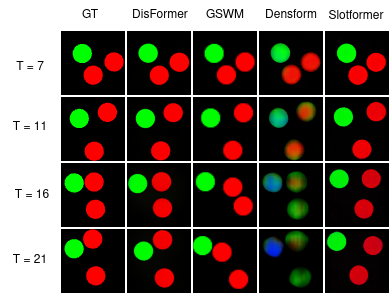}
    \caption{\label{fig:bc_ood} 2D-BC }
\end{subfigure}
~
\begin{subfigure}[t]{0.3\textwidth}
    \centering
      \includegraphics[width=4.5cm]{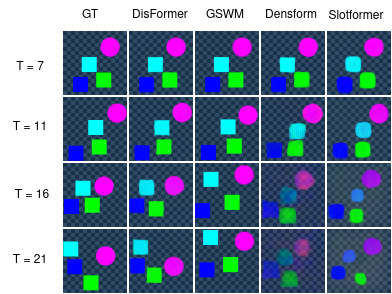}
    \caption{\label{fig:bs_ood} 2D-BS }
\end{subfigure}
~
\begin{subfigure}[t]{0.3\textwidth}
    \centering
      \includegraphics[width=4.5cm]{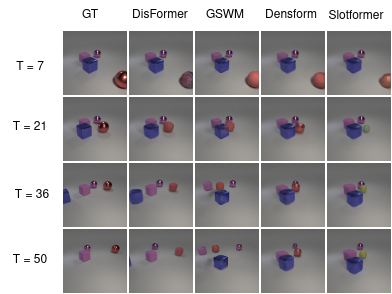}
    \caption{\label{fig:obj3d_ood}OBJ3D }
\end{subfigure}
\caption{Rollouts at various time steps for three datasets (OOD). GT: Ground Truth}
\label{fig:rollout-ood}
\end{figure*}

 {\bf 2D datasets:} Table~\ref{table:results_2d} presents the results on the 2-D datasets. \model{} beats all other approaches on both pixel-error and PSNR metrics. Our closest competitor is GSWM, and we have up to 10\% (relative) lower value of PErr on both the datasets. On PSNR, we do moderately better on 2D-BC, and marginally better on 2D-BS. Interestingly, Slotformer does quite badly on both the 2D datasets. Denseformer does somewhere in between, comparable to GSWM in PixErr and doing worse on PSNR. We see a significant gain in the OOD setting, where we beat the closest competitor GSWM, with up to relative margin of 15\% on PixErr, and a moderate gain in PSNR. There is a significant drop in the performance of Slotformer when going to OOD setting. This highlights the strength of our approach in generalizing to unseen attribute combinations. \\ 
  {\bf 3D datasets:} Table~\ref{table:results_3d} presents the results on the 3-D datasets. As in the case of 2-D datasets, \model{} beats all other approaches on both pixel-error and PSNR metrics. Our closest competitor in this case is Slotformer, which is comparable to our model in SSIM on OBJ3D, and LPIPS on  CLEVEREd, but does worse on other metrics on the two datasets. DenseFormer (on OBJ3D) performs somewhere between Slotformer and GSWM which performs worst. The strength of our approach is further highlighted in the OOD setting, where the margin of our performance gain with respect to the competitors increases significantly. We perform close to 8\% (relative) better on PSNR, 5\% (relative) better on SSIM and more than 30\% on LPIPS. Interestingly, in this case, DenseFormer is the second best performing model, beating both GSWM and Slotformer on all the merics. This setting again highlights the strenght of our models do well in OOD settiong. \\
  {\bf Visual comparisons:} Figure~\ref{fig:rollout-indist}~\ref{fig:rollout-ood} compare the predictions at various time-steps for various models in the in-distribution and OOD settings respectively, for some hand-picked examples, showing the success cases for our model compared to the baselines. The significant better quality of predictions at various values of rollouts especially in the OOD set up is clearly visible.

\begin{table*}[!ht]
\begin{minipage}[t]{1.0\textwidth}
    \centering
    \caption{Results on 2D Datasets}
    \begin{tabular}[t] {p{1.8cm}|p{0.75cm}p{0.75cm}|p{0.75cm}p{0.75cm}|p{0.75cm}p{0.75cm}|p{0.75cm}p{0.75cm}}
    \toprule
    & \multicolumn{4}{c|}{In Distribution} & \multicolumn{4}{c}{Out of Distribution} \\\hline
     &  \multicolumn{2}{c|}{2D-BC} &  \multicolumn{2}{c|}{2D-BS}  &  \multicolumn{2}{c|}{2D-BC} &  \multicolumn{2}{c}{2D-BS}\\\hline
    Model   & PErr  ($\downarrow$) & PSNR ($\uparrow$)  & PErr  ($\downarrow$)  & PSNR ($\uparrow$) & PErr  ($\downarrow$) & PSNR ($\uparrow$)   & PErr  ($\downarrow$)  & PSNR ($\uparrow$) \\\hline
    GSWM  &  0.013   &  23.2  & 0.025  & 20.0 & 0.014 & 23.0 & 0.028 & 18.51 \\
    SlotFormer  &  0.016  &  20.7 & 0.033  & 16.3 & 0.017 & 20.29 &  0.034 & 16.07\\
    DenseFormer  &  0.013   &  22.1 & 0.025 & 17.8 & 0.029 & 18.61 & 0.029 & 17.16\\
     DisFormer  &  \textbf{0.011}    & \textbf{24.3}   & \textbf{0.022}  & \textbf{20.3}  & \textbf{0.011} & \textbf{24.2} & \textbf{0.024} & \textbf{20.16} \\
    \bottomrule
    \end{tabular}
    
    \label{table:results_2d}
\end{minipage}%
~
\end{table*}
\begin{table*}[!ht]
\begin{minipage}[t]{1.0\textwidth}
    \caption{Results on 3D Datasets. For in distribution setting, the numbers for the baselines are as reported in ~\cite{slotformer}. For SSIM on OBJ3D (in distribution), our model numbers are rounded-off two to decimal places for a direct comparison.}
    \label{table:results_3d}
    \begin{tabular}[t]{p{1.8cm}|p{0.75cm}p{0.75cm}p{0.75cm}|p{0.75cm}p{0.75cm}p{0.75cm} |p{0.75cm}p{0.75cm}p{0.75cm} }
    \toprule
     & \multicolumn{6}{c|}{In Distribution} & \multicolumn{3}{c}{Out of Distribution} \\\hline
     &  \multicolumn{3}{c}{OBJ3D} &  \multicolumn{3}{c|}{CLEVRER} &  \multicolumn{3}{c}{OBJ3D} \\\cline{2-4} \cline{5-7}
    Model   & PSNR  ($\uparrow$)  & SSIM ($\uparrow$)   &  LPIPS ($\downarrow$) & PSNR  ($\uparrow$)  & SSIM ($\uparrow$)   &  LPIPS ($\downarrow$) &  PSNR  ($\uparrow$)  & SSIM ($\uparrow$)   &  LPIPS ($\downarrow$)\\\hline
    GSWM  &  31.43 & 0.89 & 0.10 & 28.42& 0.89& 0.16 & 25.75 & 0.835 &  0.155\\
    Slotformer   & 32.40  &  \textbf{0.91}  & 0.083 & 30.21 & 0.89 & \textbf{0.11} & 25.94 & 0.843 &  0.153\\
    DenFormer  &  31.9 & 0.90   & 0.091 & - & - & - & 26.69 & 0.857 & 0.13 \\
    DisFormer  &  \textbf{32.79} & \textbf{0.91}   & \textbf{0.075} & \textbf{30.36}  & \textbf{0.898} & \textbf{0.11} & \textbf{28.03}  &  \textbf{0.883}  & \textbf{0.11}\\
    \bottomrule
    \end{tabular}
    
\end{minipage}%

\end{table*}

\subsection{Disentanglement of Object Attributes}\label{subsec:disentangle}

\begin{figure*}[t]
\begin{subfigure}[t]{0.3\textwidth}
    \centering
      \includegraphics[width=4cm]{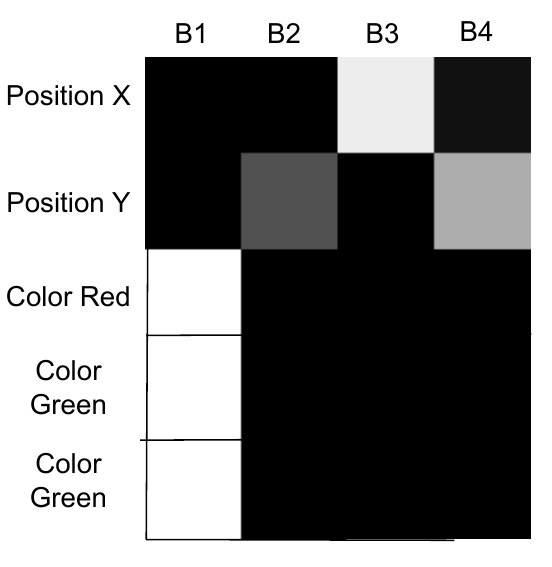}
    \caption{\label{fig:bc_imp} 2D-BC }
\end{subfigure}
~
\begin{subfigure}[t]{0.3\textwidth}
    \centering
      \includegraphics[width=4cm]{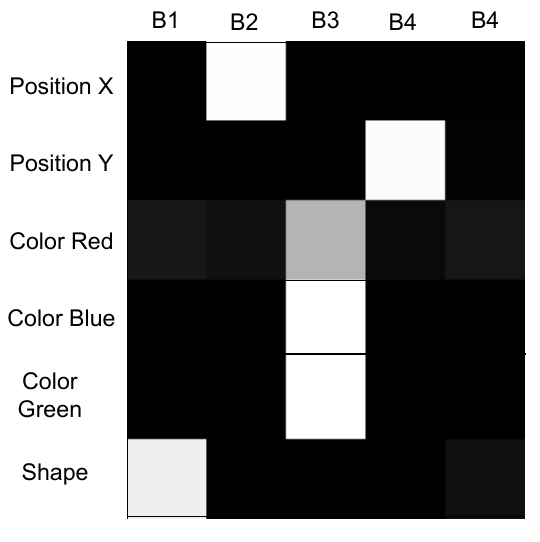}
    \caption{\label{fig:bs_1} 2D-BS }
\end{subfigure}
~
\begin{subfigure}[t]{0.3\textwidth}
    \centering
      \includegraphics[width=5cm]{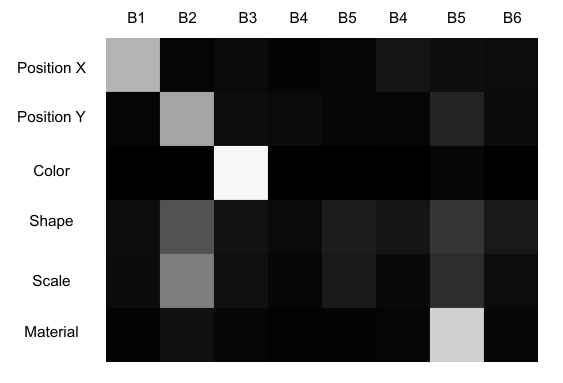}
    \caption{\label{fig:obj3d_1}OBJ3D }
\end{subfigure}
\caption{Disentanglement Results.}
\label{fig:disentanglement}
\end{figure*}

We conduct a post-hoc analysis to gain insights into the learned disentangled representations of objects in terms of their correlation to their attributes. We trained $k$ different probes corresponding to $k$ different attributes of object to predict the attribute given the block representation of object. These probes are implemented as gradient boosted trees. The ground truth attributes are generated from respective simulators. Mapping between ground truth object attributes and object blocks is done via decoder mask and simulator mask greedy matching. After training these probes we find the importance of each dimension of blocks and sum the importance of dimensions belonging to one block. This set up is very similar to the methodology adopted by ~\cite{Neural-Systematic-Binder} for evaluating the disentanglement of attributes in static images.
Figure~\ref{fig:disentanglement} shows the relative importance of each of the blocks for predicting various attributes for three of the datasets~\footnote{We did not have ground truth training masks for CLEVERER}. For 2D datasets, since each color is composed of 0-1 combination of RGB components, we create a separate probe for each of these color components. For 2D-BC, all the color probes are clustered at a single block, and other blocks taken by x coordinate, and y coordinate being represented by a combination of two blocks. This clearly points to the successful disentanglement obtained using approach. Similarly, for OBJ3D, the probes clustered across blocks, with different blocks representing material, color (RGB), X and Y, attributes respectively. There is some collusion between scale and shape attributes, pointing to the fact that model is not able fully disentangle these attributes.
\subsection{Ablation}
In this section, we analyze the robustness of our approach to variation in number of concepts and number of blocks used during training of our model. Table~\ref{table:noc_ablation} presents the performance of ~\model{} as we vary the number of concepts from 4 to 7 keeping number of blocks fixed at 6. Similarly, Table~\ref{table:nof_ablation} presents the performance as we vary the number of blocks from 6 to 8 keeping the number of concept fixed at 4. Both experiments are done in 2D-BS dataset. As can be seen, while the performance fluctuates with varying number of concepts, it goes down as the number of available blocks is increased during learning. In other words, using larger number of blocks than required result in degrading of the performance of our model. Analysing the impact of these hyperparameters more deeply is a direction for future work.
 \begin{table*}[!ht]
\begin{minipage}[t]{0.4\textwidth}
    \centering
    \caption{Number of concept ablation}
    \begin{tabular}[t] {|p{1.8cm}|p{0.75cm}p{0.75cm}|}
    \toprule
     Number of concepts ($k$) & PErr & PSNR \\
     \toprule
     4 & 0.024 & 19.2\\
     5 & 0.022 & 20.3\\
     6 & 0.023 & 19.3\\
     7 & 0.022 & 19.2\\
    \bottomrule
    \end{tabular}
    \label{table:noc_ablation}
\end{minipage}%
\hfill
\begin{minipage}[t]{0.5\textwidth}
    \centering
    \caption{Number of block ablation}
    \begin{tabular}[t] {|p{1.8cm}|p{0.75cm}p{0.75cm}|}
    \toprule
     Number of blocks ($r$) & PErr & PSNR \\
     \toprule
     5 & 0.022 & 20.3\\
     6 & 0.024 & 19.1\\
     7 & 0.027 & 16.3\\
    \bottomrule
    \end{tabular}
    \label{table:nof_ablation}
\end{minipage}%
\end{table*}

\section{Conclusion and Limitations}
\label{sec:conclusion}
We have presented an approach for learning disentangled object representation for the task of predicting visual dynamics via transformers. Our approach makes use of unsupervised object extractors, followed by learning disentangled representation by expressing dense object representation as a linear combination of learnable concept vectors. These disentangled representations are passed through a transformer to obtain future predictions. Experiments on three different datasets show that our approach performs better than existing baselines, especially in the setting of transfer. We also show that our model can indeed learn disentangled representation. Future work includes learning with more complex backgrounds, extending to more complex 3D scenes, and extending the work to action-conditioned video prediction.

{\bf Limitations:} Limitations of our work include not being able to fully disentangle the attributes for 3D datasets. Further, we have not yet performed experiments on real world datasets, so the performance of the models on these datasets remains to be seen.

\bibliographystyle{plainnat}
\bibliography{bibfile}
\newpage
\section*{A Hyperparameters} \label{apex:hyper}

\begin{table}[htb]
    \centering
    
\begin{tabular}{|c|c|c|c|c|c|}
    & &  BC & BS & OBJ3D & CLEVRER \\ \hline
    \multirow{10}{*}{Training} & Training Steps Phase 1& 30K & 30K & 60K & 80K\\
    \cline{2-6}
    & Training Steps Phase 2& 250K & 250K & 300K & 500K\\
    \cline{2-6}

    & LR warm-up steps& 1K & 1K & 1K & 1K\\
    \cline{2-6}
    & Cosine Anneal $T_{cycle}$ & 500K & 500K & 600K & 1000K \\
    \cline{2-6}
    & Cosine Anneal $T_{mult}$ & 1 & 1 &  1 & 1\\
    \cline{2-6}
    & Batch size & 32 &  32 & 32 & 32\\
    \cline{2-6}
    & $\lambda_{dec}$  & 1 & 1 & 1 & 1\\
    \cline{2-6}
    & $\lambda_{mask} $ & 0.1 & 0.1 & 0 & 0\\
    \cline{2-6}
    \hline
    \hline

    \multirow{6}{*}{Block Extractor} & Blocks $r$ & 4 & 5 & 8 & 8 \\
    \cline{2-6}
    & Concepts $k$ & 4 & 4 & 8 & 10\\
    \cline{2-6}
    & Block Dimension $d_{b}$ & 32 & 32 & 64 & 64\\
    \cline{2-6}
    & Hidden Dimension $d$ & 64 & 64 &  128 & 128\\
    \cline{2-6}
    & $T$ & 3 & 3 & 3 & 3\\
    \cline{2-6}
    & Input Dimension $d_{f}$ & 64  & 64 & 128 & 128\\

    \hline
    \hline

    \multirow{4}{*}{Transition Transformer} & Dimension $\hat{d}$ & 256 & 256 & 256 & 256\\
    \cline{2-6}
    & Layers $k$ & 4 & 4 & 4 & 4\\
    \cline{2-6}
    & Heads & 8 & 8 & 8 & 8 \\
    \cline{2-6}
    & Dropout $d$ & 0.1 & 0.1 &  0.1 & 0.1\\
    \cline{2-6}
    \hline 
    \hline
\end{tabular}
    \caption{Hyperparameters}
    \label{tab:hyperparameters}
\end{table}

\begin{table}[htb]
    \centering
    
\begin{tabular}{|c|c|c|c|c|}
    \hline
     Layers & Stride & Padding & Channels & Activation\\
      \hline
       \hline
     Conv 3x3 & 1 & 1 & 16 & ReLU \\
     \hline
     MaxPool 3x3 & 2 & 1 & - & - \\
     \hline
     Conv 3x3 & 1 & 1 & 16 & ReLU \\
     \hline
     MaxPool 3x3 & 2 & 1 & - & - \\
     \hline
     Conv 3x3 & 1 & 1 & 32 & ReLU  \\
     \hline
     MaxPool 3x3 & 2 & 1 & - & - \\
     \hline
     Conv 3x3 & 1 & 1 & 32 & ReLU \\
     \hline
     MaxPool 3x3 & 2 & 1 & - & - \\
     \hline
     Conv 3x3 & 1 & 1 & 64 & ReLU \\
     \hline
     MaxPool 3x3 & 2 & 1 & - & - \\
     \hline
     Conv 3x3 & 1 & 1 & 64 & ReLU \\
     \hline
     MaxPool 3x3 & 2 & 1 & - & - \\
     \hline
\end{tabular}
    \caption{Encoder CNN for BC, BS and OBJ3D and CLEVRER} 
    \label{tab:EncoderCNN}
\end{table}

\begin{table}[htb]
    \centering
    
\begin{tabular}{|c|c|c|c|c|c|}
    \hline
     Layers & Stride & Padding & Output Padding & Channels & Activation\\
      \hline
       \hline
     ConvTranspose 5x5 & 2 & 2 & 1 & 32 & ReLU \\
     \hline
     ConvTranspose 5x5 & 2 & 2 & 1 & 32 & ReLU \\
     \hline
     \multicolumn{6}{|c|}{Concatenate the block level output channelwise}\\
     \hline
      ConvTranspose 5x5 & 2 & 2 & 1 & 32 & ReLU \\
     \hline
     ConvTranspose 5x5 & 2 & 2 & 1 & 32 & ReLU \\
     \hline
      ConvTranspose 5x5 & 1 & 1 & 0 & 5 & ReLU \\
     \hline
     ConvTranspose 3x3 & 1 & 1 & 0 & 4 & ReLU \\
     \hline
     
\end{tabular}
    \caption{Decoder for BC, BS and OBJ3D} 
    \label{tab:2dDecoder}
\end{table}
\newpage
\section*{B} \label{apex:B}
\begin{figure*}
\centering
    \centering
    \begin{subfigure}{0.7\textwidth}
    \includegraphics[width=\linewidth]{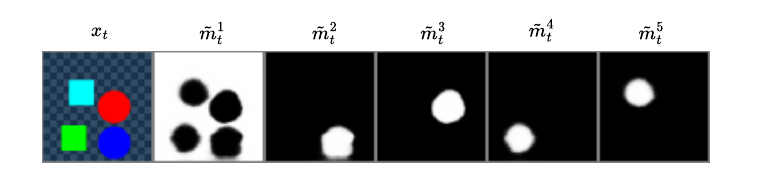}
    \caption{}
    \end{subfigure}
     \begin{subfigure}{0.7\textwidth}
    \includegraphics[width=\linewidth]{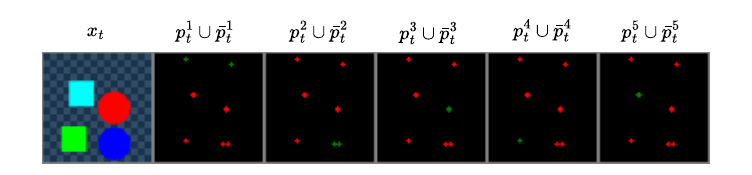}
    \caption{}
    \end{subfigure}
\begin{subfigure}{0.7\textwidth}
    \includegraphics[width=\linewidth]{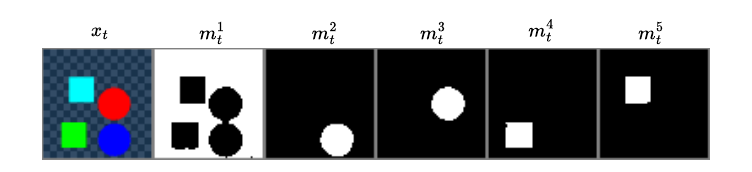}
    \caption{}
    \end{subfigure}
     
     \caption{Mask extractor: Bouncing Circles  (a) Slot masks for a frame. (b) Generated points prompts. Green points are foreground point prompts and red one are background point prompts. (c) SAM masks}
     \label{fig:obj3d_extractor}  
\end{figure*}
We describe the mask extractor in details in this section. We first train object-centric model SAVi (\underline{S}lot \underline{A}ttention for \underline{Vi}deo ) \citep{savi} for unsupervised object discovery. We use this trained object-centric model to generate prompts per object. This prompts are used to generate masks of objects from Segment SAM (Segment Anything)\citep{sam}. We now describe training of SAVi and SAM prompt generation in detail.
\subsection*{B1. SAVi Training}
Following \citep{slotformer} we pre-train SAVi to discover  objects. SAVi first applies slot attention \cite{slot_attention} on each frame to discover objects in each slots. These slots are decoded using Spatial Mixture Decoder to reconstruct the frame. The decoder is implemented as Spatial Broadcast Decoder \citep{sbd}. And for temporal consistency initializes the slots of next frame with some learnable function on current frame slots. \\
Formally given input frames $\{x_t\}_{t = 1}^T$, first each frame is passed through CNN to obtain image feature $h_{t} \in \mathbb{R}^{U \times U \times D_{in}}$. Then spatial positional encoding is added to $h_{t}$. After this per frame ordered set of $N$ slots $\mathcal{S}_t$ are obtained from $h_{t}$ by spatial competitive attention between $h_t$ and slots. For temporal consistency the slots from previous time step are used to initialize slots at current time step. This steps can be summarized as $\bar{\mathcal{S}}_{t} = f_{trans}(\mathcal{S}_{t-1})$ and $\mathcal{S}_t = f_{SA}(h_t, \bar{\mathcal{S}}_{t})$ where $f_{SA}$. Following \citep{slotformer} $f_{trans}$ is implemented as one layer transformer and $f_{SA}$ is the iterative refinement loop from \cite{slot_attention}. Each slot from ordered set $\mathcal{S}_t$ is passed through shared Spatial Broadcast Decoder to generate the mixing masks $\tilde{m}_{t}^{i}$ and content $\tilde{c}_{t}^{i}$ of Spatial Mixture Decoder. Here $i \in \{1, .., N\}$. 
We found that for 3D datasets (OBJ3D and CLEVRER) even though the slots are made compete to attend on spatial location, slot are not guaranteed to generate mutually exclusive masks $\tilde{m}_{t}^{i}$. That is two generated masks $\tilde{m}_{t}^{i}$ and $\tilde{m}_{t}^{j}$ may have high IOU score for $i \neq j$. The problem with overlapping masks is that they will generate the contradicting prompts for SAM. We further explain this in \ref{sam_prompting}. To address this issue, we found empirically that it was sufficient to add mask intersection loss as extra auxiliary loss to make masks exclusive. The mask intersection loss formally is $\mathcal{L}_{int} = \frac{2}{HWN(N-1)} \sum_{i \neq j} \sum_{a,b}^{H,W} \tilde{m}_{t}^i[a,b] \cdot \tilde{m}_{t}^j[a,b]$

\begin{figure*}
\centering
    \centering
    \begin{subfigure}{0.7\textwidth}
    \includegraphics[width=\linewidth]{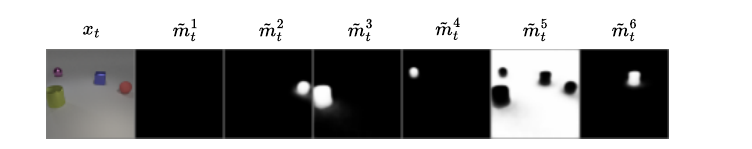}
    \caption{}
    \end{subfigure}
     \begin{subfigure}{0.7\textwidth}
    \includegraphics[width=\linewidth]{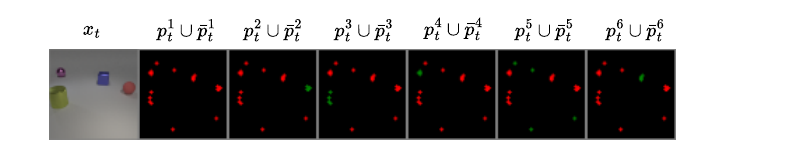}
    \caption{}
    \end{subfigure}
\begin{subfigure}{0.7\textwidth}
    \includegraphics[width=\linewidth]{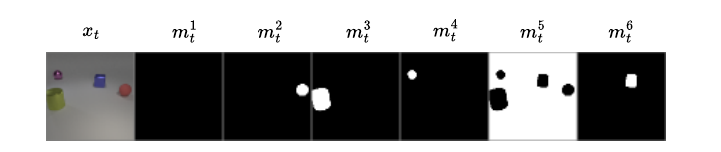}
    \caption{}
    \end{subfigure}
     \caption{Mask extractor: OBJ3D  (a) Slot masks for a frame. (b) Generated points prompts. Green points are foreground point prompts and red one are background point prompts. (c) SAM masks}
     \label{fig:extractor}  
\end{figure*}
\subsection*{B2. SAM prompting} \label{sam_prompting}
We observe that masks generated by slots, even though gives rough masks of object, can be improved. We use  Segment Anything (SAM)\cite{sam} to further improve the mask quality. SAM is promptable image segmentation model, where prompt can be foreground / background points, a approximate mask or bounding box. We generate the points and mask prompt from masks generated by slots and pass it to SAM to get better mask. \\
We generate set of $n_{p}$ foreground / positive point prompts per slot $\mathcal{S}_t^i$ denoted as $p_{t}^i$. A good choice for $p_{t}^i$ is the points which are on the object and not on the boundary of the object. Our approach is to choose pixel positions with maximum pixel values. To make sure all points are well inside the object we first threshold the slot masks $\tilde{m}_t^i$ with $thresh$ to make it binary mask $M_{t}^i$. Then convolve $M_{t}^i$ with all one filter of size $(3,3)$ for $l$ times. Denote the resultant mask as $\tilde{M}_t^i$. We choose top $n_p$ points which have maximum pixel value from $\tilde{M}_t^i$. For negative points we use the positive points of remaining slots which gives $(N-1)n_p$ negative / background points per slot. Along with this $Nn_p$ point prompt, we pass $\tilde{m}_t^i$ as mask prompt to SAM.  Algorithm \ref{alg:prompt_sampler} gives pseudo code for sampling point prompts from slot masks. \\
To handle the cases where slot does not represent object we ignore such slot masks and return all zero mask instead. If sum of pixel values of $\tilde{M}_t^i$ is less that $\texttt{mthresh}$, we consider that slot does not represent object. We use the official release of SAM with ViT-H backbone. ~\footnote{https://github.com/facebookresearch/segment-anything}. 
\begin{algorithm}
    \small{
\caption{Prompt Sampler: Inputs are: SAVi masks $\tilde{m}_{t}^i$ where $i \in \{1,..,N\}$}
\label{alg:prompt_sampler}
\begin{algorithmic}[1]
\For {$i = 1$ to $N$}
\State $M_{t}^i = m_{t}^i\texttt{.copy()}$
\State $M_t^i\texttt{[$M_t^i \leq$ thresh]} = 0$
\State $M_t^i\texttt{[$M_t^i >$ thresh]} = 1$
\For {$j = 1$ to $l$}
\State $\tilde{M}_t^i = $\texttt{Convolve(input = $M_t^i$, kernel = all\_1\_3x3, padding = same)}
\EndFor
\If {$\tilde{M}_{t}^i$.\texttt{sum()} < \texttt{mthresh}}
\State $p_{t}^i = \texttt{None}$
\Else
\State $p_{t}^i $ = \texttt{argmax\_topk($\tilde{M}_{t}^i$, k = $n_p$)}
\EndIf 
\EndFor
\State \texttt{return} $p_t^1, .., p_t^i$
\end{algorithmic}
}
\end{algorithm}
\\

\newpage
 
\newpage

\end{document}